\documentclass[acmtog]{acmart}
\usepackage{colortbl}
\usepackage{xcolor}
\usepackage{multirow}
\usepackage{float}
\usepackage{fontawesome}
\usepackage{soul}

\AtBeginDocument{%
  }

\copyrightyear{2025}
\acmYear{2025}
\setcopyright{acmlicensed}\acmConference[SA Conference Papers '25]{SIGGRAPH Asia 2025 Conference Papers}{December 15--18, 2025}{Hong Kong, Hong Kong}
\acmBooktitle{SIGGRAPH Asia 2025 Conference Papers (SA Conference Papers '25), December 15--18, 2025, Hong Kong, Hong Kong}
\acmDOI{10.1145/3757377.3763990}
\acmISBN{979-8-4007-2137-3/2025/12}

\begin{document}

\title{CamPVG: Camera-Controlled Panoramic Video Generation with Epipolar-Aware Diffusion}

\author{Chenhao Ji}
\email{jichenhao@tongji.edu.cn}
\authornote{Work done during an internship in DAMO Academy, Alibaba Group.}
\orcid{0009-0005-6718-4582}
\affiliation{%
  \institution{Tongji University}
  \city{Shanghai}
  \country{China}
}
\affiliation{%
  \institution{DAMO Academy, Alibaba Group}
  \city{Hangzhou}
  \country{China}
}

\author{Chaohui Yu}
\email{huakun.ych@alibaba-inc.com}
\orcid{0000-0002-7852-4491}
\affiliation{%
  \institution{DAMO Academy, Alibaba Group}
  \city{Beijing}
  \country{China}
}

\author{Junyao Gao}
\email{junyaogao@tongji.edu.cn}
\orcid{0009-0004-6527-8166}
\affiliation{%
  \institution{Tongji University}
  \city{Shanghai}
  \country{China}
}

\author{Fan Wang}
\email{fan.w@alibaba-inc.com}
\orcid{0000-0001-7320-1119}
\affiliation{%
  \institution{DAMO Academy, Alibaba Group}
  \city{Hangzhou}
  \country{China}
}

\author{Cairong Zhao}
\authornote{Corresponding author.}
\email{zhaocairong@tongji.edu.cn}
\orcid{0000-0001-6745-9674}
\affiliation{%
  \institution{Tongji University}
  \city{Shanghai}
  \country{China}
}

\begin{teaserfigure}
  \vspace{10pt}
  \includegraphics[width=\textwidth]{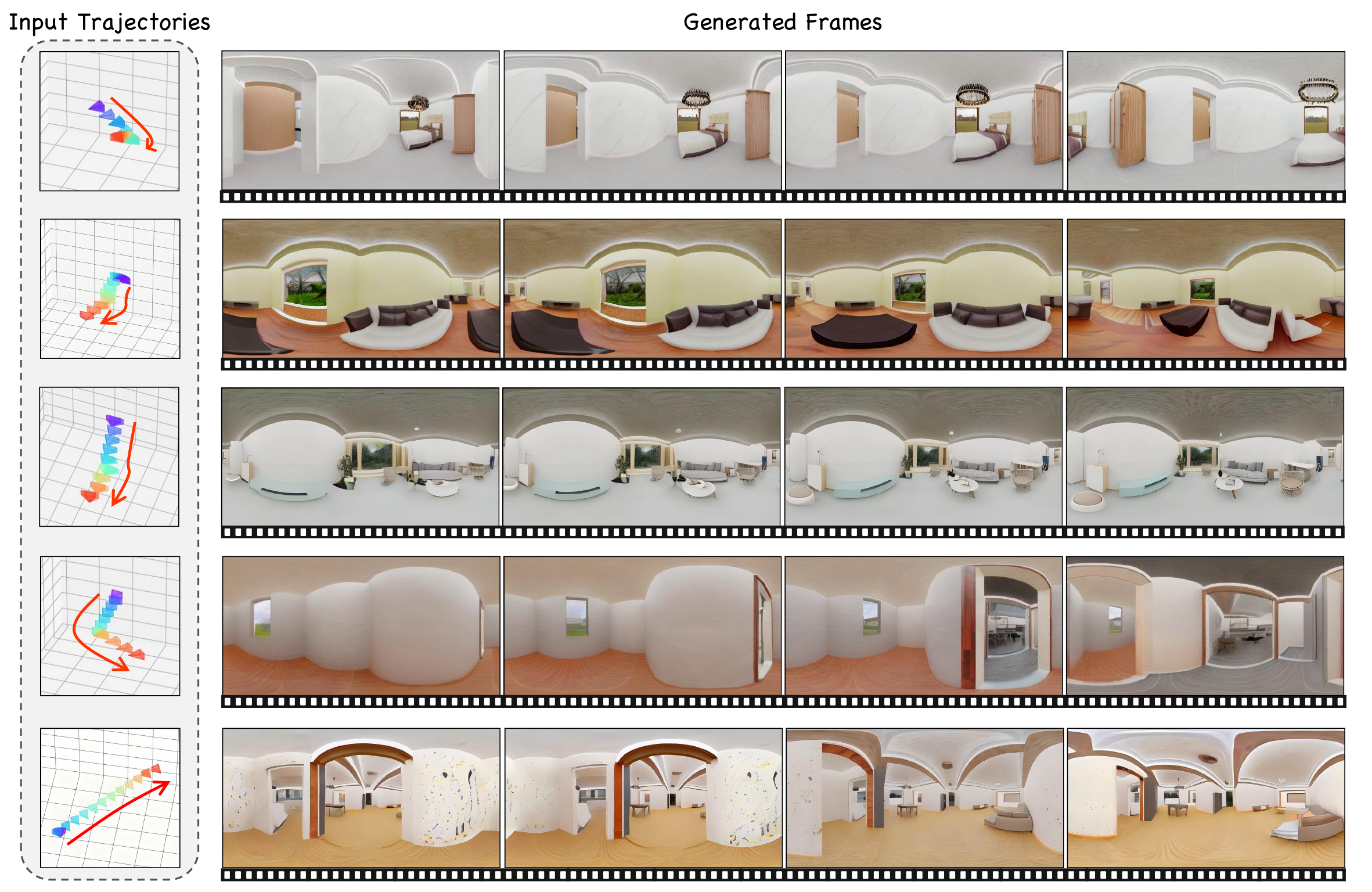}
  \caption{\textbf{CamPVG} is the first camera-controlled panoramic video generation framework. Given a specified camera trajectory and an initial conditional frame, it generates high-quality panoramic videos with strong geometric consistency across the panoramic space. By leveraging panoramic Plücker embeddings and a spherical epipolar-aware module, our method effectively models global geometric structures and viewpoint transitions, generating spatially coherent and visually realistic panoramic videos.}
  \Description{Teaser.}
  \label{fig:teaser}
  \vspace{10pt}
\end{teaserfigure}

\setlength{\abovecaptionskip}{5pt}
\setlength{\belowcaptionskip}{-8pt}
\definecolor{highlightcolor}{RGB}{233,247,217}

\begin{abstract}
Recently, camera-controlled video generation has seen rapid development, offering more precise control over video generation.
However, existing methods predominantly focus on camera control in perspective projection video generation, while geometrically consistent panoramic video generation remains challenging.
This limitation is primarily due to the inherent complexities in panoramic pose representation and spherical projection.
To address this issue, we propose CamPVG, the first diffusion-based framework for panoramic video generation guided by precise camera poses.
We achieve camera position encoding for panoramic images and cross-view feature aggregation based on spherical projection.
Specifically, we propose a panoramic Plücker embedding that encodes camera extrinsic parameters through spherical coordinate transformation.
This pose encoder effectively captures panoramic geometry, overcoming the limitations of traditional methods when applied to equirectangular projections.
Additionally, we introduce a spherical epipolar module that enforces geometric constraints through adaptive attention masking along epipolar lines.
This module enables fine-grained cross-view feature aggregation, substantially enhancing the quality and consistency of generated panoramic videos.
Extensive experiments demonstrate that our method generates high-quality panoramic videos consistent with camera trajectories, far surpassing existing methods in panoramic video generation.
\end{abstract}



\maketitle

\section{Introduction}
The rapid advancement of virtual reality (VR), metaverse technologies, and embodied artificial intelligence has catalyzed a surge of interest in panoramic visual content.
Panoramic videos, which capture a comprehensive 360-degree view of the surrounding environment, offer users an immersive experience. 
As the applications of panoramic videos expand, the demand for enhanced immersive experiences has continually increased.
The development of scalable panoramic video generation with precise camera pose control has emerged as a critical area of research.
This capability opens new frontiers for expansive applications in entertainment, interaction, and beyond.

Generating panoramic videos with consistent camera motion and temporal smoothness poses distinctive challenges.
Many existing works on panoramic video generation, such as 360DVD~\cite{360DVD}, Imagine 360~\cite{imagine360}, and 4K4DGen~\cite{4k4dgen}, primarily focus on generating dynamic content in panoramic videos but offer limited control over camera perspectives. 
Consequently, the generated panoramic videos exhibit minimal variation in viewpoints.
To achieve precise camera pose control, recent approaches have made notable strides. 
For instance, MotionCtrl~\cite{MotionCtrl} concatenates camera poses with features in the latent space, while CameraCtrl~\cite{CameraCtrl} injects camera poses into the latent space using Plücker embedding.
In addition, CamCo~\cite{CamCo} and CamI2V~\cite{CamI2V} further enhance scene consistency across different camera viewpoints by introducing geometric constraints through epipolar attention.
However, these methods are specifically designed for perspective projection video generation and show limitations when applied to panoramic domain.
These limitations stem primarily from the challenging representation of panoramic camera poses and the intrinsic geometric complexity of panoramic imagery.

To address these challenges, we propose CamPVG, the first diffusion-based framework for panoramic video generation guided by precise camera poses.
Our approach enables the generation of high-quality panoramic videos that maintain consistency with given camera trajectories as shown in Fig.~\ref{fig:teaser}.
Unlike prior perspective-based geometric methods, our approach is not simply an adaptation of perspective models to panoramic coordinate systems.
Existing camera pose encoding methods are typically designed for perspective projection camera trajectories and perform poorly on equirectangular panoramic data due to the inherent differences in imaging logic between panoramic and perspective views.
To achieve effective camera position encoding for panoramic data, we introduce \textbf{panoramic Plücker embedding}, building upon the foundation of traditional Plücker embedding~\cite{CameraCtrl}. 
Our method models the spatial relationship between each pixel in the panoramic image and the camera origin through spherical projection, representing this relationship using Plücker coordinates~\cite{pluck}.
These spherically projected Plücker coordinates are then injected into the latent space via a pose encoder, providing spatial geometric guidance throughout the panoramic video generation process.
Furthermore, to enhance consistency across different camera viewpoints, we propose a \textbf{spherical epipolar module} for fine-grained feature aggregation. 
By leveraging the intrinsic properties of equirectangular projection, we calculate the spherical epipolar lines corresponding to each pixel across different viewpoints.
We then employ spherical epipolar masking with carefully designed sampling strategy along the epipolar line to filter out irrelevant pixel information.
During the panoramic video generation process, we aggregate valid reference information from different viewpoints using spherical epipolar attention, thereby achieving multi-view consistent panoramic video generation.

Extensive experiments demonstrate that CamPVG achieves superior performance in camera trajectory consistency, frame realism, and overall video quality. 
Our method significantly surpasses existing camera-controlled video generation approaches in panoramic video generation.
We believe that CamPVG will make substantial contributions to the field of camera pose-guided panoramic video generation and its downstream applications. 
Our contributions can be summarized as follows:
\begin{itemize}
    \vspace{-5pt}
    \item We propose CamPVG, the first framework for panoramic video generation guided by precise camera poses, enabling the generation of high-quality panoramic videos with consistent camera trajectories.
    \item We introduce panoramic Plücker embedding, a novel approach for camera position encoding based on panoramic data.
    \item We present the spherical epipolar module that leverages spherical epipolar constraints to achieve fine-grained feature aggregation, enhancing multi-view consistency and visual fidelity of panoramic videos.
\end{itemize}

\begin{figure*}
    \centering
    \includegraphics[width=\textwidth]{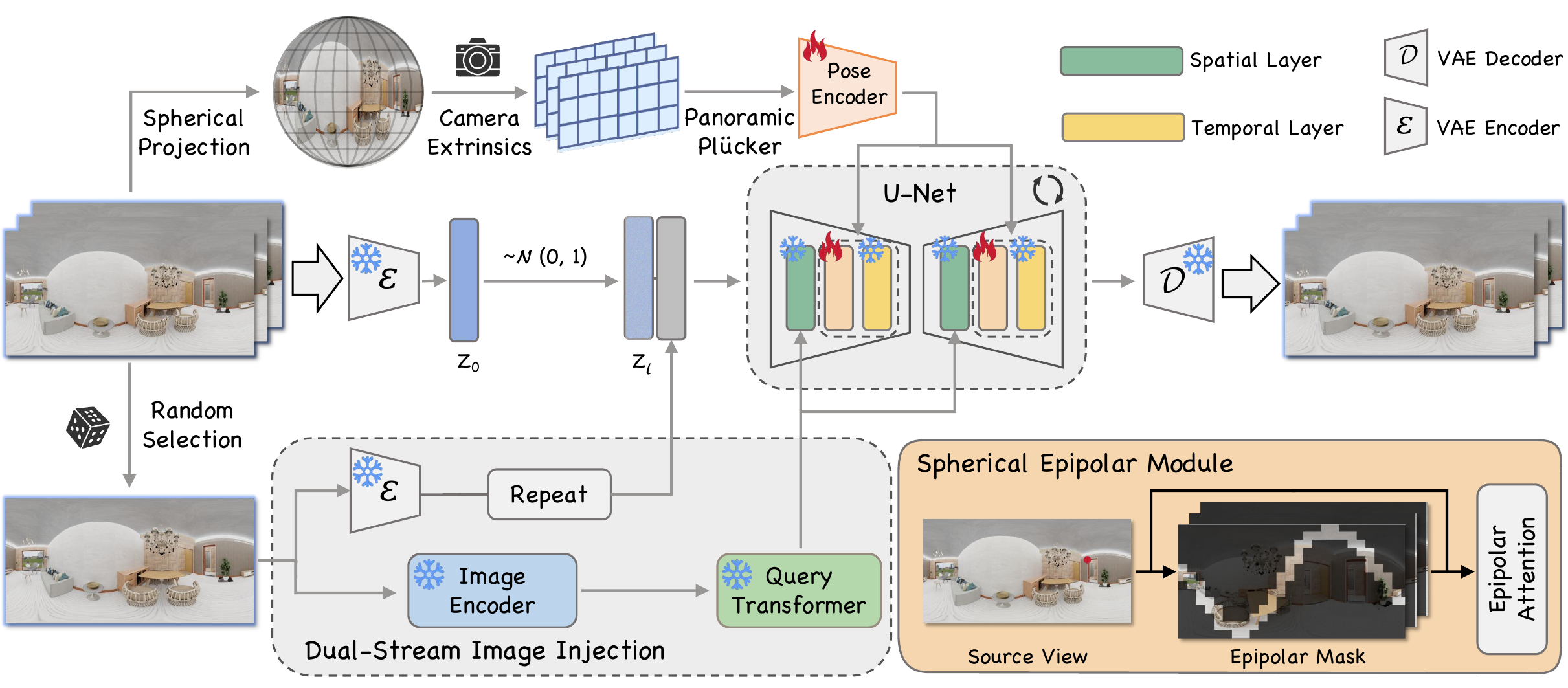}
    \caption{\textbf{Framework of CamPVG.} CamPVG employs spherical projection to transform input camera trajectories into panoramic Plücker embeddings, which are injected into the U-Net to guide panoramic geometry learning. Additionally, the spherical epipolar module computes epipolar masks through cross-view geometric constraints and applies spherical epipolar attention to enhance multi-view consistency. This integrated approach enables precise camera-controlled panoramic video generation.}
    \Description{framework}
    \label{fig:framework}
\end{figure*}

\section{Related Work}
\subsection{Diffusion-Based Video Generation}
Recent advancements in diffusion models \cite{gao2024styleshot, gao2025charactershot,gao2025faceshot,tang2025lego,jiang2024delving,jiang2024global} have significantly advanced research in video generation.
Building on the success of text-to-image (T2I) diffusion frameworks such as Stable Diffusion~\cite{stable_diffusion}, researchers have extended these models to text-to-video (T2V) generation by incorporating temporal layers to process video input while retaining strong visual priors.
For instance, Video Diffusion Model~\cite{video_diffusion_model}, LVDM~\cite{lvdm}, and VideoCrafter~\cite{VideoCrafter1,VideoCrafter2} extend the 2D U-Net architecture of image diffusion models with spatial and temporal blocks, enabling coherent video generation through iterative denoising.
Furthermore, Sora~\cite{sora} and CogVideoX~\cite{CogVideoX} explore Transfomer-based diffusion framework integrated with 3D-VAE, significantly enhanced the video generation capabilities in terms of temporal consistency and visual fidelity.
Additionally, other works~\cite{svd,DynamiCrafter} have advanced image-to-video (I2V) generation by conditioning diffusion models on image inputs.

\subsection{Panoramic Video Generation}
The field of panoramic content generation has seen notable progress with the advent of diffusion models, though most efforts remain focused on static panorama synthesis~\cite{mvdiffusion,diffpano,panfusion,CamFreeDiff,Pathdreamer} rather than video generation.
Recent studies~\cite{DynamicScaler,VideoPanda} have increasingly explored diffusion-based frameworks to overcome these limitations.
For instance, 360DVD~\cite{360DVD} introduces a lightweight 360-Adapter to fine-tune pre-trained T2I diffusion models, enabling panoramic video synthesis conditioned on textual prompts and motion signals.
Imagine360~\cite{imagine360} proposes a dual-branch architecture that enforces joint local and global constraints, facilitating perspective-to-panoramic video conversion.
Another approach, 4K4DGen~\cite{4k4dgen}, leverages 2D priors from perspective image generation models to denoise spherical latent codes, yet the generated videos suffer from restricted viewpoint diversity due to inadequate motion modeling in the latent space.
While these methods demonstrate promising progress, their controllability over camera trajectories remains limited. 
Our CamPVG advances the field by integrating explicit camera pose conditioning into the diffusion framework, enabling precise control over viewpoint transitions in panoramic video generation.

\subsection{Camera-Controlled Video Generation}
Camera-controlled video generation has emerged as a critical research direction in diffusion-based video generation, aiming to produce dynamic visual content aligned with predefined camera trajectories.
Some approaches~\cite{MotionMaster,Peekaboo} achieve coarse camera motion control through training-free methods.
To enable precise camera control, recent works integrate camera pose information into diffusion frameworks.
MotionCtrl~\cite{MotionCtrl} concatenates noisy latent features with camera pose in temporal blocks, allowing camera-conditioned generation.
Similarly, CameraCtrl~\cite{CameraCtrl} encodes Plücker embeddings~\cite{pluck} and injects them into the U-Net architecture.
While these methods demonstrate improved controllability, their ability to model 3D spatial relationships still limits.
Addressing this limitation, some methods~\cite{CamCo,CamI2V,Collaborative_Video_Diffusion} introduce epipolar attention mechanisms to explicitly model 3D geometric constraints.
While effective for perspective-view generation, their effectiveness remains constrained in panoramic video generation.
Our work extends this principle to panoramic domains by reformulating epipolar attention for equirectangular projections.

\section{Method}
In this section, we introduce our novel method for panoramic video generation guided by precise camera poses with spherical epipolar constraints, as illustrated in Fig.~\ref{fig:framework}. 
We begin with the preliminary concepts of controllable video diffusion models and the representation of camera poses in Sec.~\ref{sec:pre}.
To encode camera trajectories for panoramas, we propose a panoramic Plücker embedding in Sec.~\ref{sec:pano_cam}.
To better capture geometric constraints between multi-view panoramic frames, Sec.~\ref{sec:epi} details the proposed spherical epipolar module.

\subsection{Preliminary}
\label{sec:pre}
\subsubsection{Controllable Video Diffusion Model}
Modern diffusion-based video generation frameworks synthesize content guided by multi-modal conditional inputs.
These models enable user-specified video synthesis by conditioning the generation process on diverse signals, including textual prompts, reference image, and motion information.
The framework operates in a compressed latent space derived through a learned auto-encoder architecture.
Given an input video sequence $x \in \mathbb{R}^{N\times H\times W\times 3}$ comprising $N$ frames of resolution $H \times W$, the encoder $\mathcal{E}$ produces latent representations $z_{0}^{1: N}=\mathcal{E}$.
During training, Gaussian noise $\epsilon \sim \mathcal{N}(\mathbf{0}, \mathbf{I})$ is progressively added across $t$ diffusion steps, producing the noised latent $z_{t}^{1: N}$.
The denoising model $\epsilon_\theta$ learns to predict the noise $\epsilon$ conditioned on the input signals at the time step $t$.
The training objective can be formulated as follows:
\begin{equation}
\mathcal{L}=\mathbb{E}_{\mathcal{E}(x), \epsilon 
\sim \mathcal{N}(\mathbf{0}, \mathbf{I}), c_{t}, t}\left[\left\|\epsilon-\hat{\epsilon}_{\theta}\left(z_{t}^{1: N}, c_{t}, t\right)\right\|_{2}^{2}\right],
\end{equation}
where $c_{t}$ represents the embeddings of conditional information.
This formulation enables joint optimization of spatial-temporal coherence and conditional alignment across modalities.

\subsubsection{Camera Representation}
The pose of the camera is defined by both intrinsic and extrinsic parameters.
The intrinsic $\mathbf{K}\in\mathbb{R}^{3\times3}$ establishes the mapping from the camera coordinate system to the pixel coordinate system.
The extrinsic $\mathbf{E}\in\mathbb{R}^{3\times4}$, which includes a rotation matrix $\mathbf{R}\in$ SO(3) and a translation vector $\mathbf{t}\in\mathbb{R}^{3}$, specifies the camera’s orientation and position in the world coordinate system.
Alternatively, camera poses can be encoded through Plücker embeddings~\cite{pluck}, which parameterize the relationship between image pixels and 3D rays originating from the camera.
For each pixel $\left( u,v \right)$, its Plücker embedding $\mathbf{P}_{u,v}=\left(\mathbf{m},\mathbf{d}\right)\in\mathbb{R}^{6}$ is defined as follows:
\begin{itemize}
    \item $\mathbf{d}\in\mathbb{R}^{3}$ represents the direction of the 3D ray from the camera center to the pixel in world coordinates. 
    \item $\mathbf{m}\in\mathbb{R}^{3}$
    denotes the moment vector, computed as the cross product between the position of the camera center and the direction vector $\mathbf{d}$. 
\end{itemize}
Given camera-to-world extrinsic $\mathbf{E}=[\mathbf{R},\mathbf{t}]$ and intrinsic $\mathbf{K}$, the Plücker embedding for pixel $\left( u,v \right)$ is derived via:
\begin{equation}
\label{pluck_compute}
\mathbf{d} = \mathbf{R} \left( \mathbf{K}^{-1} \left(u,v,1\right)^\top \right), \quad
\mathbf{m} = \mathbf{t} \times \mathbf{d}.
\end{equation}

\begin{figure*}
    \centering
    \setlength{\abovecaptionskip}{10pt}
    \includegraphics[width=\textwidth]{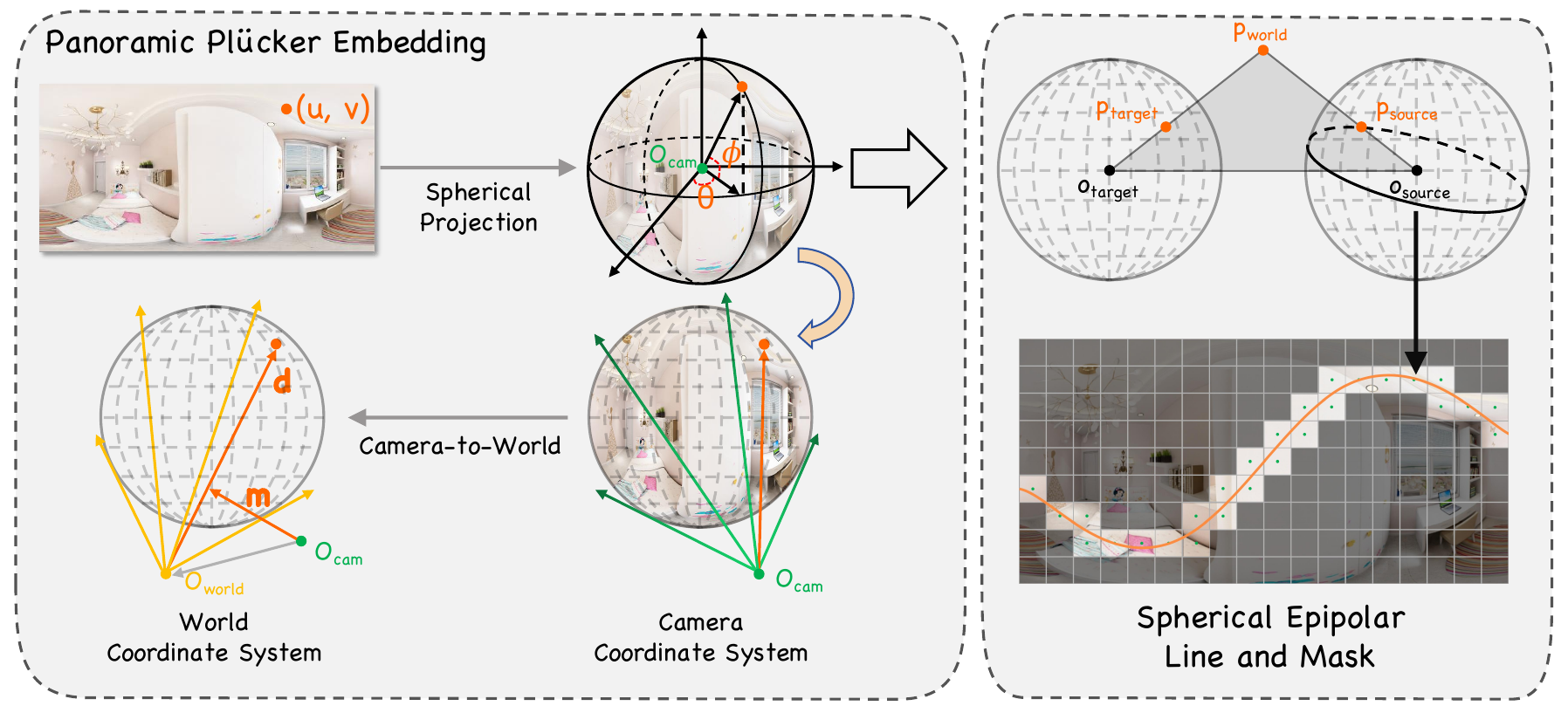}
    \caption{\textbf{Panoramic Plücker Embedding and Epipolar Geometry.} Left: transformation from pixel coordinates to Panoramic Plücker embedding. Right: epipolar geometry relationship of pixel points across different coordinate systems.}
    \label{fig:epi}
    \Description{spherical module}
\end{figure*}
\setlength{\abovecaptionskip}{5pt}

\vspace{-10pt}
\subsection{Panoramic Camera Pose Representation}
\label{sec:pano_cam}
Directly incorporating camera extrinsic parameters into video diffusion for learning viewpoint transitions poses significant challenges.
To address this, we leverage Plücker embeddings that explicitly model the geometric relationship between pixels and camera rays.
Compared with raw extrinsics, Plücker embeddings offer better numerical stability and richer geometric cues through their uniform magnitude distribution across the scene.
However, conventional Plücker computation assumes perspective projection with known intrinsic parameters, which are undefined for equirectangular panoramas.
For panoramas represented in equirectangular projection, we derive a spherical direction vector for each pixel via spherical projection and then calculate the corresponding panoramic plunker embedding, as illustrated in Fig.~\ref{fig:epi} (Left). 
Given a pixel $\left( u,v \right)$ in a panorama with resolution $H\times W$, its corresponding spherical coordinates are computed as:
\begin{equation}
\label{pix2sphere}
\begin{aligned}
\phi = \frac{u}{W}\cdot 2\pi,\quad
\theta =\frac{v}{H}\cdot \pi.
\end{aligned}
\end{equation}
Through the calculated azimuth $\phi$ and elevation $\theta$ angles in spherical coordinates, we can transform these into directional vectors in the Cartesian coordinate system:
\begin{equation}
\label{sphere2cas}
\begin{aligned}
x_{\left(u,v\right)} =\cos(\theta) \cdot \sin(\phi),\;
y_{\left(u,v\right)} =\sin(\theta),\;
z_{\left(u,v\right)} =\cos(\theta) \cdot \cos(\phi).
\end{aligned}
\end{equation}
This spherical-to-cartesian conversion establishes a consistent 3D position mapping for panorama pixels, enabling Plücker embedding computation without conventional camera intrinsics.
According to Eq.(\ref{pluck_compute}), the Plücker embeddings for each panorama pixel with extrinsic $\mathbf{E}=[\mathbf{R},\mathbf{t}]$ are computed as:
\begin{equation}
\mathbf{d} = \mathbf{R}\left(\hat{x}_{\left(u,v\right)},\hat{y}_{\left(u,v\right)},\hat{z}_{\left(u,v\right)}\right)^\top , \quad
\mathbf{m} = \mathbf{t} \times \mathbf{d},
\end{equation}
where $\left(\hat{x}_{\left(u,v\right)},\hat{y}_{\left(u,v\right)},\hat{z}_{\left(u,v\right)}\right)^\top$ represents the normalized direction vector.
We construct the complete camera trajectory $\mathbf{P}\in\mathbb{R}^{N\times H\times W\times 6}$ by converting each panoramic video frame's camera extrinsics into Plücker embeddings.
Following CameraCtrl~\cite{CameraCtrl}, we employ a trainable pose encoder with linear projection layer to map the trajectory into latent representations.
These encoded camera features are subsequently integrated into the diffusion U-Net to enable camera-aware generation.

\subsection{Spherical Epipolar Module}
\label{sec:epi}
\subsubsection{Spherical Epipolar Line}
Epipolar geometry establishes geometric constraints for potential pixel correspondences across multi-view images.
In perspective projection, epipolar lines can be directly computed through the essential matrix derived from relative camera poses and intrinsic parameters, resulting in straight lines in planar images.
For equirectangular panoramas, however, the equirectangular projection necessitates a modified approach to epipolar geometry due to the non-linear coordinate mapping, as shown in Fig.~\ref{fig:epi} (Right).
Given two panoramic views with extrinsics $\left[\mathbf{R}_{i}, \mathbf{t}_{i}\right]$ and $\left[\mathbf{R}_{j}, \mathbf{t}_{j}\right]$, we compute their relative pose as:
\begin{equation}
\mathbf{R}_{i\to j}=\mathbf{R}_j\cdot \mathbf{R}_i^{-1}, \quad 
\mathbf{t}_{i\to j}=\mathbf{t}_j-\mathbf{R}_{i\to j}\mathbf{t}_i.
\end{equation}
A pixel $\left(u_i, v_i\right)$ in view $i$ converts to 3D Cartesian coordinates $\mathbf{p}_i=\left(x_i, y_i, z_i\right)$ through Eq.(\ref{pix2sphere}) and Eq.(\ref{sphere2cas}).
The corresponding projected point in view $j$ becomes $\mathbf{p}_{i\to j} = \mathbf{R}_{i\to j} \cdot \mathbf{p}_i + \mathbf{t}_{i\to j}$.
Similarly, the camera origin $\mathbf{o}_i$ projects to view $j$ as $\mathbf{o}_{i\to j}=\mathbf{t}_{i\to j}$.
The spherical epipolar line of point $\mathbf{p}_i$ comprises projections of points along ray $\overrightarrow{\mathbf{o}_i\mathbf{p}_i}$, which lie on the plane $\Pi$ containing $\mathbf{o}_j$, $\mathbf{p}_{i\to j}$, and $\mathbf{o}_{i\to j}$. 
For equirectangular projection, the intersection of plane $\Pi$ with the coordinate sphere in spherical coordinates represents the corresponding epipolar line.
Expressing the plane $\Pi$ in the camera coordinate system of view $j$ as $Ax+By+Cz+D=0$, we derive coefficients through geometric constraints: 
\begin{equation}
    \begin{aligned}
        A&=\frac{z_{\mathbf{o}_{i\to j}}\cdot y_{\mathbf{p}_{i\to j}}-y_{\mathbf{o}_{i\to j}} \cdot z_{\mathbf{p}_{i\to j}}}{y_{\mathbf{o}_{i\to j}}\cdot x_{\mathbf{p}_{i\to j}}-x_{\mathbf{o}_{i\to j}} \cdot y_{\mathbf{p}_{i\to j}}} \cdot C=A'\cdot C, \\
        B&=\frac{z_{\mathbf{o}_{i\to j}} \cdot x_{\mathbf{p}_{i\to j}}-x_{\mathbf{o}_{i\to j}} \cdot z_{\mathbf{p}_{i\to j}}}{x_{\mathbf{o}_{i\to j}}\cdot y_{\mathbf{p}_{i\to j}} -y_{\mathbf{o}_{i\to j}} \cdot x_{\mathbf{p}_{i\to j}}} \cdot C = B'\cdot C, \\
        D&= 0,
    \end{aligned}
\end{equation}
where $\left(x_{\mathbf{o}_{i\to j}}, y_{\mathbf{o}_{i\to j}},z_{\mathbf{o}_{i\to j}} \right)$ and $\left(x_{\mathbf{p}_{i\to j}}, y_{\mathbf{p}_{i\to j}},z_{\mathbf{p}_{i\to j}} \right)$ denote coordinates of $\mathbf{o}_{i\to j}$ and $\mathbf{p}_{i\to j}$ respectively.
Combining with the spherical constraint in Eq.(\ref{sphere2cas}) and converting to pixel coordinates via Eq.(\ref{pix2sphere}), we obtain the epipolar line parametrization:
\begin{equation}
v = -\frac{H}{\pi} \left( {\arctan  \frac{A' \sin  \frac{2 \pi u}{W}  + \cos  \frac{2 \pi u}{W} }{B'}} \right),
\label{eq:epipolar}
\end{equation}
where $\left(u,v\right)$ represents pixel coordinates in view $j$'s panorama.

\subsubsection{Spherical Epipolar Mask}
\label{sec:k_points}
The epipolar line defines geometrically valid correspondences between source and target views by establishing plausible reference pixels.
While perspective projection enables efficient distance computation through linear epipolar constraints, spherical geometry requires non-linear treatment due to curved epipolar trajectories derived from Eq.(\ref{eq:epipolar}).
We compute the minimum spherical distance between a pixel $\mathbf{p}$ and the epipolar line through discretized samples along the curve.
To mitigate computational complexity, we uniformly sample $K$ points $\left \{ \mathbf{c}_k\right \}_{k=1}^K $ along the epipolar line and approximate the minimum distance as:
\begin{equation}
d_{\text{min}} = \min_{1\leq k\leq K} \|\mathbf{p} - \mathbf{c}_k\|_2.
\end{equation}
A pixel qualifies as a valid reference when $d_{\text{min}}$ falls below half the feature grid's diagonal length. 
This thresholding strategy ensures geometrically consistent correspondences while accommodating localization uncertainties.
For each panoramic frame $i$ in the video sequence, we compute per-pixel epipolar masks across all frames through spherical geometry constraints.
The complete spherical epipolar mask $\mathbf{M}_i\in\mathbb{R}^{HW\times N \times HW}$ is obtained by aggregating these view-consistent correspondences.
As visualized in Fig.~\ref{fig:epi} (Right), the resultant binary mask restricts cross-view attention to topologically aligned regions while preserving multi-view consistency.

\begin{table*}[t!]
  \setlength{\belowcaptionskip}{0pt}
  \caption{\textbf{Quantitative Comparisons with Baseline Methods.} Our method demonstrates significant improvements across three critical dimensions compared to baseline methods: camera view consistency, photorealistic fidelity of generated frames, and holistic video quality.}
  \label{tab:quantity}
  \centering
  \resizebox{1.0\linewidth}{!}{ %
  \begin{tabular}{l|ccc|c|cc|ccc}
    \toprule
    \multirow{2}*{\bf{Method}}   & 
    \multirow{2}*{\bf{LPIPS$\downarrow$}}  & \multirow{2}*{\bf{SSIM$\uparrow$}}  & \multirow{2}*{\bf{PSNR$\uparrow$}}  &
    \multirow{2}*{\bf{FAED$\downarrow$}}  & \multicolumn{2}{c}{\bf{FVD}$\downarrow$} &
    \multicolumn{3}{c}{\bf{VBench}$\uparrow$}\\
    ~&~&~&~&~& VideoGPT & StyleGAN 
    &Aesthetic Quality & Subject Consistency & Temporal Flickering\\
    \midrule
    MotionCtrl~\cite{MotionCtrl}
    &0.1741 &0.6008 &29.51 &0.2993 &94.84 &73.75 &0.4878 &0.8706 &0.9256  \\
    CameraCtrl~\cite{CameraCtrl}
    &0.1816 &0.5926 &29.29 &0.3967 &151.08 &135.90 &0.4832 &0.8593 &0.9241\\
    CamI2V~\cite{CamI2V}
    &0.1867 &0.5887 &29.24 &0.2621 &91.83 &76.25 &0.4931 &0.8755 &0.9302  \\
    \rowcolor{highlightcolor}\bf{CamPVG (ours)} &\bf{0.1480} &\bf{0.6544} &\bf{30.05} &\bf{0.1066} &\bf{66.24} &\bf{56.34} &\bf{0.5043} &\bf{0.9000} &\bf{0.9339} \\
    \bottomrule
  \end{tabular}}
  \vspace{-10pt}
\end{table*}
\setlength{\belowcaptionskip}{-8pt}

\begin{figure}[ht]
    \vspace{-5pt}
    \centering
    \setlength{\abovecaptionskip}{2pt}
    \setlength{\belowcaptionskip}{-5pt}
    \includegraphics[width=\linewidth]
    {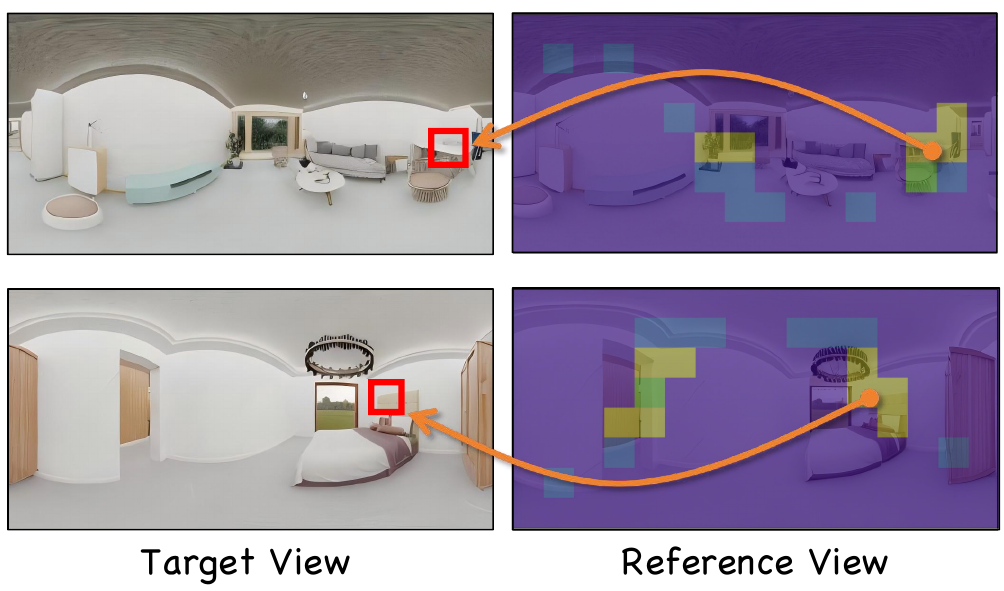}
    \caption{\textbf{Visualization Results of Spherical Epipolar Attention Map.}}
    \label{fig:epi_attention}
    \Description{qualitative_comparison}    
\end{figure}
\setlength{\abovecaptionskip}{5pt}
\setlength{\belowcaptionskip}{-8pt}

\subsubsection{Spherical Epipolar Attention}
We introduce spherical epipolar attention to enforce multi-view consistency in panoramic video generation through explicit geometric constraints. 
In video diffusion models, spatial attention primarily focuses on the spatial relationships within single-frame images, while temporal attention mainly addresses the relationships between consecutive frames.
Hence, we apply spherical epipolar attention before the temporal attention to facilitate the model's learning of correspondences between different viewpoints.
For each query frame $q_i\in \mathbb{R}^{HW\times C}$, the key and value are derived from all $N$ frames as $k\in \mathbb{R}^{NHW\times C}$ and $v\in \mathbb{R}^{NHW\times C}$.
The attention computation incorporates our precomputed spherical epipolar mask $\mathbf{M}_i\in\mathbb{R}^{HW\times N \times HW}$as:
\begin{equation}
\text{SphericEpiAttn}(q_i, k, v) = \text{softmax}\left(\frac{q_ik^\top}{\sqrt{d}} \odot \mathbf{M}_i\right)v,
\end{equation}
where $d$ represents the dimension of attention heads. 
The visualization results of spherical epipolar attention map are shown in Fig.~\ref{fig:epi_attention}. This architectural modification enables simultaneous learning of temporal dynamics and cross-view geometric relationships, particularly crucial for maintaining 3D consistency during panoramic camera motion.
The explicit geometric prior embedded in the attention mechanism guides the diffusion model to preserve scene structure across viewpoints without requiring explicit 3D reconstruction.

\section{Experiments}

\begin{figure*}
    \centering
    \includegraphics[width=\textwidth]{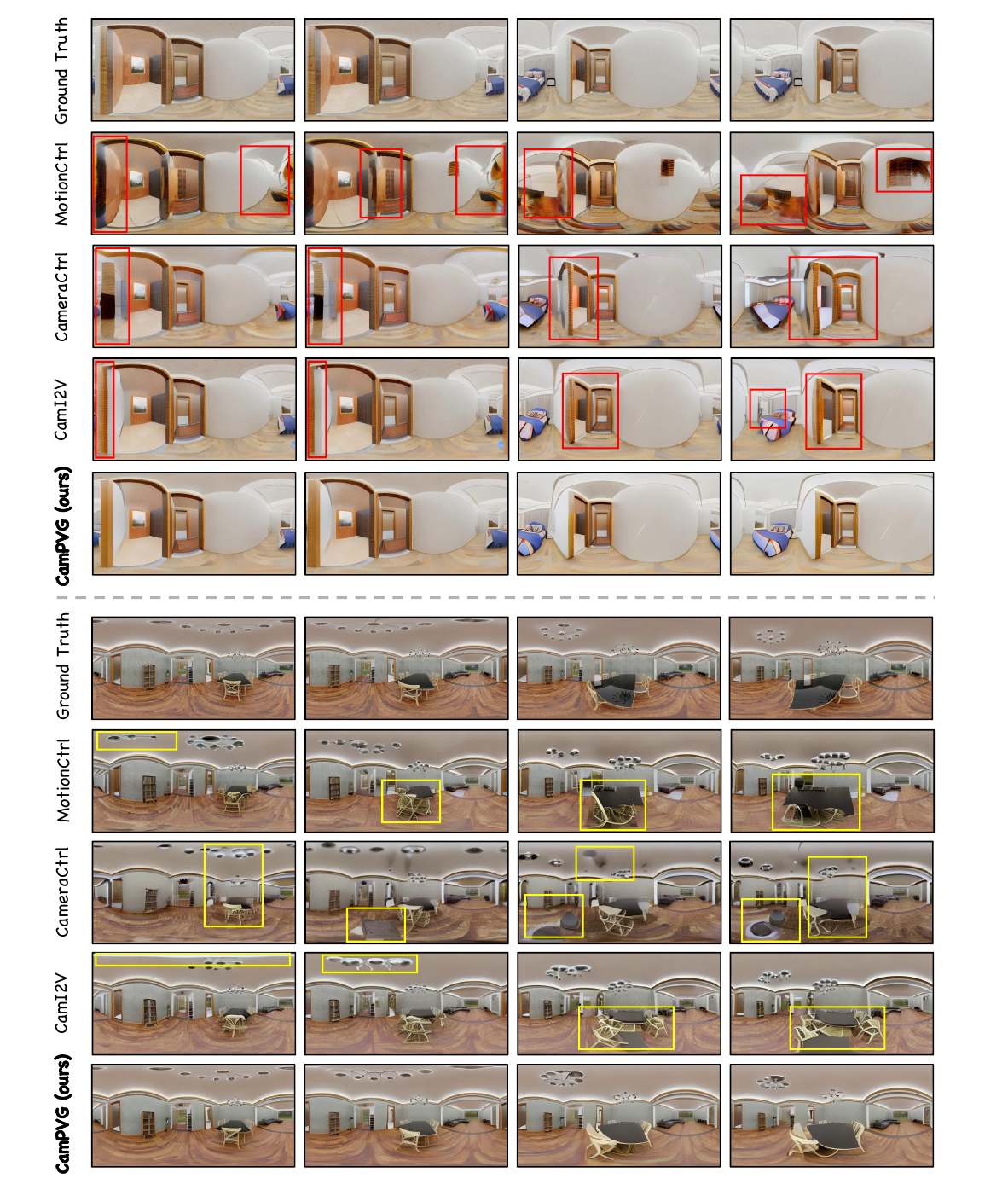}
    \caption{\textbf{Qualitative Comparison with Baseline Methods.}}
    \label{fig:qualitative}
    \Description{qualitative_comparison}
\end{figure*}

\begin{figure*}
    \centering
    \includegraphics[width=\textwidth]{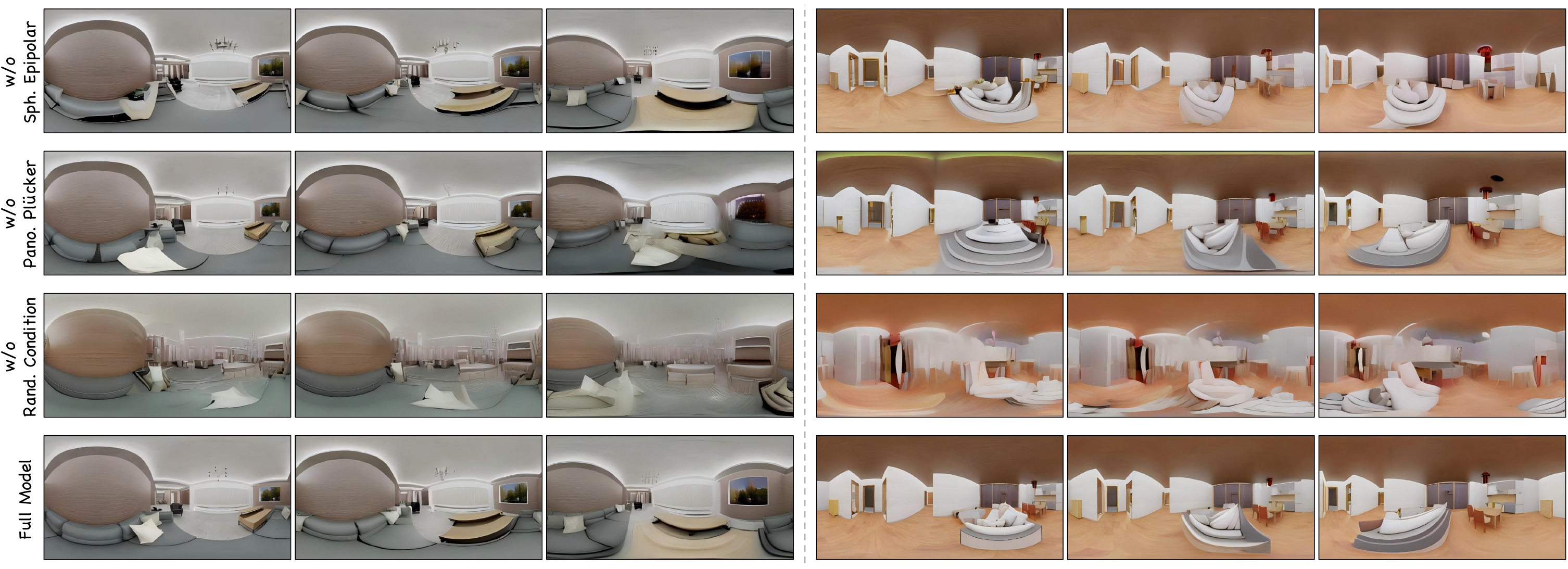}
    \caption{\textbf{Qualitative Ablation Study on Different Model Components.} Removing any individual component leads to a noticeable degradation in visual quality and temporal coherence, whereas the complete model consistently achieves the best overall performance.}
    \label{fig:visual_ablation}
    \Description{visual ablation}
    \vspace{15pt}
\end{figure*}

\begin{table*}[ht]
  \caption{\textbf{Ablation Study on Different Components.} We perform an ablation study evaluating the impact of removing the panoramic Plücker embedding, spherical epipolar module, and random conditional frame strategy on model performance.}
  \label{tab:ablation_module}
  \centering
  \begin{tabular}{l|ccc|ccc|c|cc}
    \toprule
    \multirow{2}*{\bf{Method}}   &
    \multirow{2}*{\bf{\shortstack{Pano.\\Plücker}}} &
    \multirow{2}*{\bf{\shortstack{Sph.\\Epipolar}}} &
    \multirow{2}*{\bf{\shortstack{Rand.\\Condition}}}&
    \multirow{2}*{\bf{LPIPS$\downarrow$}}  & \multirow{2}*{\bf{SSIM$\uparrow$}}  & \multirow{2}*{\bf{PSNR$\uparrow$}}  &
    \multirow{2}*{\bf{FAED$\downarrow$}}  & \multicolumn{2}{c}{\bf{FVD}$\downarrow$} \\
    ~&~&~&~&~&~&~&~&VideoGPT &StyleGAN \\
    \midrule
    \multirow{4}*{\shortstack{CamPVG\\(ours)}}
    &\faCheck &\faCheck &\faCheck &\bf{0.1480} &\bf{0.6544} &\bf{30.05} &\bf{0.1066} &\bf{66.24} &\bf{56.34} \\
    ~&\faCheck & &\faCheck &0.1865 &0.5998 &29.48 &0.2704 &87.14 &77.13 \\
    ~ & &\faCheck &\faCheck &0.3278 &0.4868 &28.85 &0.4954 &124.93 &101.03 \\
    ~&\faCheck &\faCheck & &0.4144 &0.4794 &28.51 &0.7591 &259.03 &157.92\\
    \bottomrule
  \end{tabular}
  \vspace{-5pt}
\end{table*}

\subsection{Experiment Settings}
\subsubsection{Dataset.}
To obtain panoramic video datasets incorporating precise camera poses, we construct camera trajectories within the 3D-FRONT dataset~\cite{3d-front}.
At each position along the trajectory, we render cubemaps and convert them into panoramas through equirectangular projection.
We render panoramic videos for 5,616 scenes within 3D-FRONT.
We generate 40-frame sequences for each camera trajectory, and randomly sample 16 frames at a resolution of $256\times512$ to form individual video clips.
Following CamI2V~\cite{CamI2V}, we implement randomized conditional frame selection as data augmentation.

\vspace{-3pt}
\subsubsection{Implementation Details.}
We choose DynamiCrafter~\cite{DynamiCrafter} as our base image-to-video model, removing its text conditioning component and generating 16-frame panoramic videos.
During training, we freeze all the parameters of the base model and only train our panoramic position encoder and spherical epipolar module.
The spherical epipolar mask computation samples $K=250$ points along each epipolar line for distance approximation.
We employ the Adam optimizer with a fixed learning rate of $1\times 10^{-4}$.
The model is trained on $8\times$ NVIDIA A800 GPUs with a batch size of 16 for 300 epochs, taking approximately 4 days to complete.
For fair comparison, we retrain baseline methods on our panoramic dataset with the same training settings. 

\vspace{-3pt}
\subsubsection{Evaluation Metrics.}
As conventional pose estimation methods~\cite{colmap,glomap} for perspective videos are inapplicable to panoramic content, we assess camera trajectory consistency by comparing the fifth generated frame after conditioning inputs with corresponding ground truth frame. 
This temporal offset allows us to avoid minimal differences in adjacent frames and excessive generative divergence in distant frames.
Frame-wise similarity is evaluated through Peak Signal-to-Noise Ratio (PSNR), Structural Similarity Index (SSIM)~\cite{ssim}, and Learned Perceptual Image Patch Similarity (LPIPS)~\cite{lpips}. 
To evaluate the visual fidelity of generated panoramic frames, we compute the Fréchet Auto-Encoder Distance (FAED)~\cite{panfusion} on selected fifth-frame instances.
FAED extends the Fréchet Inception Distance (FID)~\cite{FID}, and is specifically designed to address equirectangular projection distortions.
Additionally, we evaluate the overall quality of the panoramic video using Fréchet Video Distance (FVD)~\cite{fvd,VideoGPT,StyleGAN} and VBench~\cite{VBench}. 
All metrics are computed over 1,000 randomly sampled video clips.

\subsection{Comparisons with Baseline Methods}

\subsubsection{Quantitative Comparisons}
As we propose the first framework for precise camera pose-guided panoramic video generation, existing methods are not directly comparable. 
Consequently, we adapt and retrain three perspective-domain approaches, including CameraCtrl~\cite{CameraCtrl}, MotionCtrl~\cite{MotionCtrl}, and CamI2V~\cite{CamI2V}. 
For fair comparison, we modify MotionCtrl by retaining only its camera control module while disabling object motion components. 
All methods utilize DynamiCrafter as the base model and are trained on our panoramic video datasets with precise camera pose annotations.
As demonstrated in Tab.~\ref{tab:quantity}, CamPVG achieves superior performance in camera view consistency metrics (PSNR, SSIM and LPIPS), indicating more accurate reconstruction of panoramic views with less distortion and higher structural consistency.. 
Furthermore, our method preserves high video generation quality and visual fidelity, achieving the lowest FVD and FAED scores and the highest VBench scores, owing to the incorporation of geometric constraints.
These results demonstrate that our geometric-aware constraints bridge panoramic consistency and generation fidelity.

\subsubsection{Qualitative Comparisons}
\begin{figure*}[!t]
    \centering
    \includegraphics[width=\textwidth]{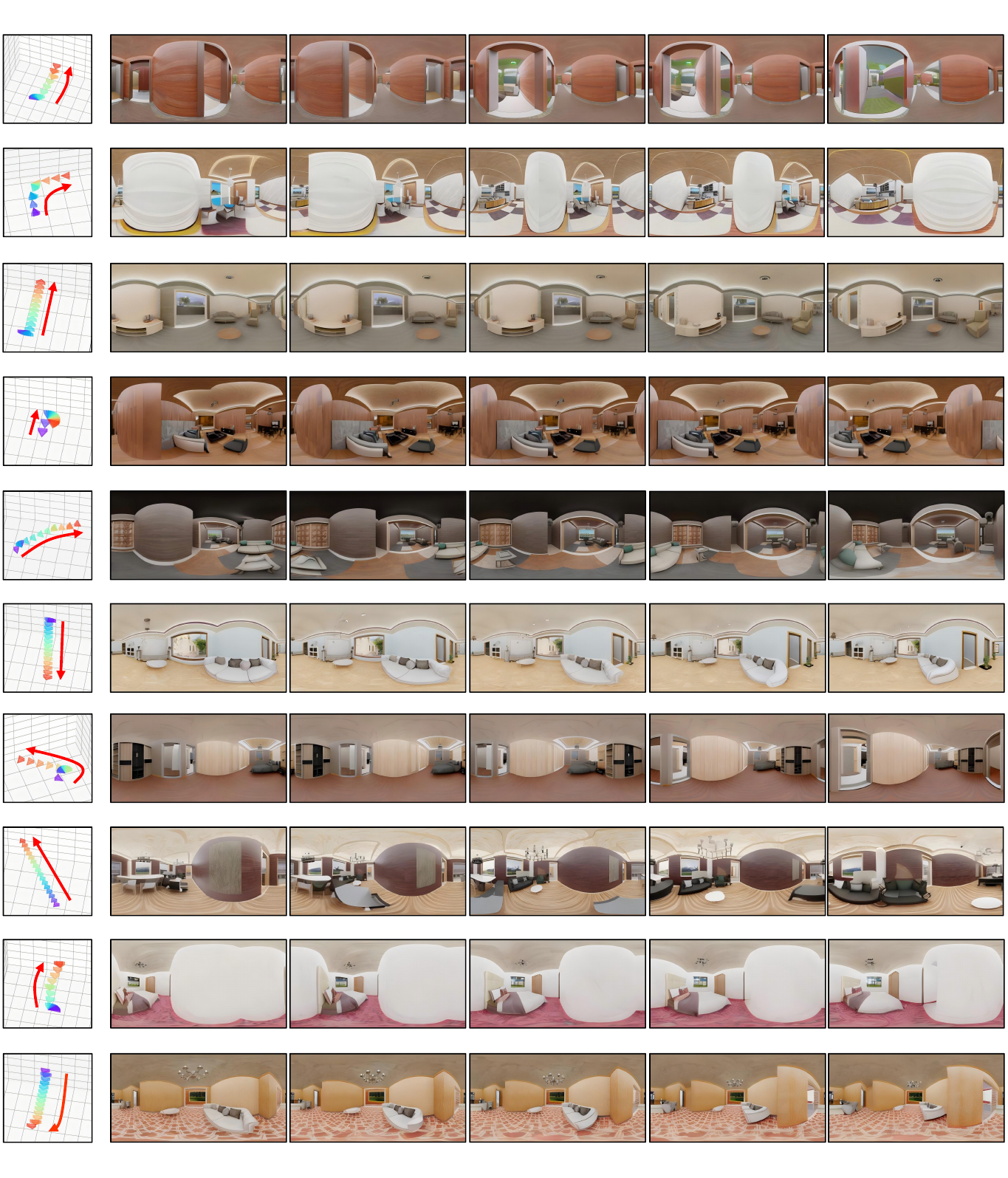}
    \caption{\textbf{More Generated Results of CamPVG.}}
    \label{fig:more_results}
    \Description{qualitative_comparison}
\end{figure*}

We present a qualitative comparison between our method and existing baseline approaches in Fig.~\ref{fig:qualitative}.
MotionCtrl~\cite{MotionCtrl} controls camera viewpoints by concatenating camera poses directly with latent space features.
While this approach achieves reasonable results in perspective video generation, it fails to model panoramic spatial geometry, leading to content loss and inconsistent cross-view alignment in panoramic scenarios. 
CameraCtrl~\cite{CameraCtrl} incorporates camera poses into the U-Net via perspective-based Plücker embeddings.
However, its perspective-centric positional encoding cannot address panoramic geometric distortions, resulting in noticeable content deformation across frames. 
CamI2V~\cite{CamI2V} further introduces a perspective-projection-based epipolar module. 
However, due to the domain gap between perspective and panoramic representations, its epipolar module fails to correctly compute the corresponding epipolar lines for target viewpoints, leading to degraded feature referencing.
As a result, CamI2V struggles to preserve fine-grained details and occasionally produces content deformations.
Especially in complex scenes (e.g., the second example in Fig.~\ref{fig:qualitative}), it exhibits cross-view inconsistency and generates hallucinated content due to the absence of panoramic geometric priors.
In contrast, our method explicitly models panoramic camera poses and enforces spherical epipolar constraints, ensuring high-fidelity geometric consistency across dynamically changing viewpoints. 
The panoramic videos generated by our approach effectively mitigate distortions while preserving intricate scene details.
Even in complex scenarios with multi-room transitions, our approach maintains strict alignment with the input camera trajectory without introducing unrealistic artifacts.
More generated results are shown in Fig.~\ref{fig:more_results}.

\begin{table*}[ht]
  \setlength{\belowcaptionskip}{0pt}
  \caption{\textbf{Ablation Study on Sampling Density.} Both insufficient and excessive numbers of sampling points degrade model performance, with the best performance observed when $K=250$.}
  \label{tab:ablation_k}
  \centering
  \begin{tabular}{l|ccc|c|cc}
    \toprule
    \multirow{2}*{\bf{Number of $K$}}   &
    \multirow{2}*{\bf{LPIPS$\downarrow$}}  & \multirow{2}*{\bf{SSIM$\uparrow$}}  & \multirow{2}*{\bf{PSNR$\uparrow$}}  &
    \multirow{2}*{\bf{FAED$\downarrow$}}  & \multicolumn{2}{c}{\bf{FVD}$\downarrow$}\\
    ~&~&~&~&~&VideoGPT  & StyleGAN \\
    \midrule
    $K=100$
    &0.1525 &0.6398 &29.91 &0.1240 &72.68 &63.78 \\
    $K=150$
    &0.1499 &0.6429 &29.98 &0.1146 &71.22 &63.56 \\
    $K=200$
    &0.1500 &0.6457 &29.94 &0.1141 &67.66 &57.63  \\
    \rowcolor{highlightcolor}$K=250$
    &\bf{0.1480} &\bf{0.6544} &\bf{30.05} &\bf{0.1066} &\bf{66.24} &\bf{56.34}  \\
    $K=300$
    &0.1486 &0.6458 &29.99 &0.1130 &75.43 &68.37 \\
    \bottomrule
  \end{tabular}
\end{table*}

\begin{table*}[ht]
  \vspace{5pt}
  \caption{\textbf{User Study.} More participants prefer the panoramic videos generated by our CamPVG. The right two columns show the comparison results under real-world inputs. Our method achieves higher preference rates across all metrics.}
  \label{tab:user_study}
  \centering
  \begin{tabular}{l|ccc|cc}
    \toprule
    \multirow{2}*{\bf{Method}}&
    \multirow{2}*{\bf{\shortstack{Camera\\Consistency$\uparrow$}}} & \multirow{2}*{\bf{\shortstack{Condition\\Consistency$\uparrow$}}}&
    \multirow{2}*{\bf{\shortstack{Video\\Quality$\uparrow$}}}&
    \multirow{2}*{\bf{\shortstack{Condition\\Consistency$\uparrow$}}}&
    \multirow{2}*{\bf{\shortstack{Video\\Quality$\uparrow$}}}\\
    ~&~&~&~&~&\\
    \midrule
    MotionCtrl~\cite{MotionCtrl} &1.895 &1.898 &1.788 &1.483 &1.500\\
    CameraCtrl~\cite{CameraCtrl} &1.835 &1.743 &1.895 &1.927 &1.877\\
    CamI2V~\cite{CamI2V} &2.753 &2.763 &2.770 &2.750 &2.850\\
    \rowcolor{highlightcolor} \textbf{CamPVG (ours)}
    &\bf{3.518} &\bf{3.598} &\bf{3.548} &\bf{3.817} &\bf{3.783}\\
    \bottomrule
  \end{tabular}
  \vspace{-5pt}
\end{table*}

\subsection{Ablation Study}
\subsubsection{Ablation Study on Different Components.}
Our method integrates geometric constraints through the panoramic Plücker embedding and the spherical epipolar module. 
Additionally, we employ a random conditional frame strategy during training to enhance the model's robustness.
To validate the contribution of each component within CamPVG, we conduct ablation studies focusing on these three critical components.
Qualitative (Fig.~\ref{fig:visual_ablation}) and  quantitative results (Tab.\protect~\ref{tab:ablation_module}) are presented.
The results show that removing the panoramic Plücker embedding leads to a larger performance drop than removing the spherical epipolar module.
The panoramic Plücker embedding enables the model to comprehend panoramic camera pose information; without it, the model loses the ability to represent the panoramic space, leading to substantial performance degradation.
In contrast, the spherical epipolar module reinforces geometric constraints across different viewpoints, thereby enhancing fine-grained consistency between generated frames. 
These findings indicate that the panoramic Plücker embedding is fundamental for establishing global camera pose awareness, while the spherical epipolar module complements it by ensuring cross-view geometric consistency.
Additionally, to assess the impact of the random conditional frame strategy, we evaluate models trained without it by adopting a different conditional frame order during testing.
As shown in the results, models trained with a fixed frame order suffer a significant performance drop across all evaluation metrics.
This degradation is caused by overfitting to the specific order observed during training, which limits the model’s ability to adapt when the order is altered at inference time.
These results demonstrate that the random conditional frame strategy is crucial for enhancing both the robustness and generalization capability of the model.

\subsubsection{Ablation Study on Sampling Density.}
As discussed in Sec.~\ref{sec:k_points}, spherical epipolar lines are characterized by uniformly sampling points along the curve. 
The number of sampling points significantly impacts model performance.
Insufficient sampling may miss valid correspondences due to large intervals between samples, while excessive sampling introduces noise from invalid references. 
To determine the optimal sampling density, we conduct ablation experiments with $K\in\{100, 150, 200, 250, 300\}$, while keeping the width of the generated videos at 512 pixels.
As shown in Tab.~\ref{tab:ablation_k}, the model achieves the best results in camera view consistency, photorealistic fidelity, and overall video quality when $K=250$.
Moreover, increasing the number of reference points to $K=300$ leads to performance degradation due to an excess of irrelevant points. 
These findings emphasize the necessity of carefully selecting an appropriate $K$ to optimize the model's ability to capture essential geometric details.

\subsection{User Study}
To complement our quantitative comparison, we conduct a human evaluation to compare our CamPVG against baseline methods (MotionCtrl, CameraCtrl, and CamI2V). 
For each method, we generate 20 panoramic video sequences using identical conditional frames.
We invited 20 volunteers to evaluate the generated panoramic videos across three dimensions: camera trajectory consistency, consistency with the conditional images, and overall video quality.
Participants rate each aspect on a scale from 1 to 4, with higher scores indicating better performance.
To further evaluate the generalization capability of our method, we additionally select real-world panoramic images as conditional inputs and conduct an extended user study.
As shown in Tab.~\ref{tab:user_study}, CamPVG achieves superior ratings across all evaluation dimensions, consistently outperforming the baseline methods.
Notably, even under real-world inputs, our method attains the highest performance (see the rightmost two columns of the table). 
These results highlight the effectiveness of CamPVG.
Additional qualitative results in diverse scenarios are provided in the supplementary material.

\section{Conclusion}
In this work, we propose CamPVG, the first diffusion-based framework for panoramic video generation guided by precise camera poses. 
By introducing panoramic Plücker embedding with the pose encoder, our method effectively learns panoramic camera geometry, enabling more accurate modeling of camera trajectories based on panoramic images.
Additionally, through the spherical epipolar module, we achieve fine-grained feature aggregation by leveraging features along epipolar lines from different viewpoints, thereby enhancing consistency across video frames.
Compared to other camera-controlled video generation methods, our approach demonstrate state-of-the-art performance in panoramic video generation, excelling in camera trajectory consistency, frame realism, and overall video quality.

\paragraph{Limitation}
Our method is currently limited by the availability of panoramic datasets with accurate camera pose annotations, which affects its performance in complex outdoor environments. 
Due to these dataset constraints, the generated panoramic videos are primarily from static scenes. 

\paragraph{Future Work}
In future work, we plan to incorporate dynamic-scene panoramic data to enhance motion realism and improve the generalization ability of our framework.

\begin{acks}
This work was supported by National Natural Science Fund of China (No.62473286).
\end{acks}
    
\bibliographystyle{ACM-Reference-Format}
\bibliography{CamPVG}

\end{document}